\documentclass[letterpaper]{article} 
\usepackage{aaai2026}  
\usepackage{times}  
\usepackage{helvet}  
\usepackage{courier}  
\usepackage[hyphens]{url}  
\usepackage{graphicx} 
\usepackage{natbib}  
\usepackage{caption} 
\usepackage{algorithmic}

\usepackage{newfloat}
\usepackage{listings}
\usepackage{booktabs}
\usepackage{amssymb}
\usepackage{amsmath}
\usepackage{tkz-graph}
\usepackage[procnumbered,ruled,linesnumbered,vlined ]{algorithm2e} 
\newcounter{algoline}
\newcommand{\nlast}{\refstepcounter{algoline}\nlset{\rlap{\textsuperscript{*}}}}
\SetKwInput{KwInput}{Inputs}
\SetKwInput{KwOutput}{Output}
\SetKwFor{ForEach}{for each}{do}{end}
\usepackage{pgfplots}
\pgfplotsset{compat=newest}
\DeclareCaptionStyle{ruled}{labelfont=normalfont,labelsep=colon,strut=off} 
\title{A Fast Heuristic Search Approach for Energy-Optimal Profile Routing for Electric Vehicles}
\author{Saman Ahmadi,
Mahdi Jalili}
\affiliations{
School of Engineering, RMIT University, Australia\\
    saman.ahmadi@rmit.edu.au, mahdi.jalili@rmit.edu.au
}

\usepackage{bibentry}

\begin{document}

\maketitle

\begin{abstract}
We study the energy-optimal shortest path problem for electric vehicles (EVs) in large-scale road networks, where recuperated energy along downhill segments introduces negative energy costs. While traditional point-to-point pathfinding algorithms for EVs assume a known initial energy level, many real-world scenarios involving uncertainty in available energy require planning optimal paths for all possible initial energy levels, a task known as energy-optimal profile search. Existing solutions typically rely on specialized profile-merging procedures within a label-correcting framework that results in searching over complex profiles.
In this paper, we propose a simple yet effective label-setting approach based on multi-objective A* search, which employs a novel profile dominance rule to avoid generating and handling complex profiles. We develop four variants of our method and evaluate them on real-world road networks enriched with realistic energy consumption data. Experimental results demonstrate that our energy profile A* search achieves performance comparable to energy-optimal A* with a known initial energy level.
\end{abstract}

\section{Introduction}
Optimal pathfinding is a well-known and classic problem in AI: determining the least-cost path from a starting point to a target location within a network.
For networks with non-negative edge costs, this problem can be efficiently solved using dynamic programming techniques such as Dijkstra’s algorithm \cite{dijkstra1959note}, or more efficient heuristic-guided methods like A* search \cite{hart1968formal}.
However, the problem becomes more challenging in the presence of negative edge costs.
In this case, traditional algorithms like Bellman-Ford \cite{bellman1958routing,ford1956network} are applicable but tend to be slower than their counterparts designed for graphs with non-negative costs.
Nonetheless, if the graph is free of negative cycles, more efficient techniques such as Johnson’s algorithm \cite{johnson1977efficient} can be used. These methods achieve performance comparable to Dijkstra’s algorithm, albeit after a reweighting step.
A practical application of optimal pathfinding with negative edge costs arises when energy is the primary cost metric, as it can be both consumed (positive cost) and regenerated (negative cost) along an edge.
Due to the law of conservation of energy, such graphs are typically free of negative cycles, making reweighting techniques viable in practice.

One key example of optimal pathfinding with energy costs is path planning for electric vehicles (EVs) in transportation networks.
Driven by the rapid advancement of vehicle and battery technologies, EVs have permeated nearly every form of transportation, including (but not limited to) daily commuting with light EVs, public transit with electric buses, and package delivery with e-bikes.
In the context of pathfinding, EVs exemplify systems with negative energy costs, as they feature energy recuperation (e.g., via regenerative braking), which can result in negative energy costs on certain road segments.
Given that the energy available to an EV is constrained by its battery capacity, reaching certain destinations (including charging stations) may only be possible via an energy-optimal path, i.e., a path from origin to destination with the lowest possible energy requirement.
Energy-optimal paths are critical as they determine feasibility: an EV cannot reach its destination if no minimum-energy path exists for the available energy at the origin.
It is therefore essential to have algorithms capable of reasoning about and efficiently computing energy-optimal paths.

When the EV's energy at the starting point is known, energy-optimal pathfinding can be formulated as a single-criteria pathfinding problem, where the available energy at any node along the energy-optimum path to the destination can be computed through simple algorithmic modifications to ensure it remains within the range.
The work in \citet{artmeier2010shortest} represents one of the first attempts to solve this problem efficiently by such modification through two approaches: an adapted version of the Bellman-Ford algorithm (with polynomial time complexity), and a modified version of Dijkstra’s algorithm that allows vertex re-exploration (with exponential time complexity). 
This research was later extended in \citet{SachenbacherLAH11}, where the authors applied Johnson’s reweighting technique \cite{johnson1977efficient} and an adapted A*-based algorithm to further reduce computational complexity to quadratic time.
A key component of their algorithms is a potential energy function that enables the upshifting of edge costs into non-negative reduced costs, and also serves as a lower-bound heuristic in the A* framework.
Inspired by this idea, subsequent research has focused on preprocessing-based approaches to accelerate energy-optimum path queries. These methods include applying Johnson’s edge-cost shifting technique to precompute a graph with only non-negative edge costs, as well as hierarchical speed-up techniques such as Contraction Hierarchies (CH) \cite{geisberger2008contraction}.
In particular, the work in \citet{baum2013energy, DBLP:journals/transci/BaumDGWZ19} continues this line of research, exploring various potential functions for cost shifting and evaluating their integration with speed-up techniques such as CH \cite{DellingGPW17, JungP02}.

A more general case of energy-optimal pathfinding arises when the initial energy level is unknown. This critical scenario appears in many real-world applications, such as planning multi-leg trips in EV routing, where the initial energy level of each leg depends on the final state of charge (SoC) from the previous trip \cite{AhmadiTHK21_cp}, or in bidirectional settings (with a backward profile search from destination to origin), where the energy level at the destination cannot be predetermined \cite{DBLP:journals/algorithmica/BaumDPSWZ20}.
In this setting, cost-optimal paths must be computed for all possible energy values, ranging from zero to EV's maximum battery capacity. This task is known as \textit{energy profile search}.
Surprisingly, energy-optimal profile search for EVs remains under-explored. The first attempts to address this problem were made by \citet{baum2013energy} and \citet{SchonfelderLW14}, where Dijkstra’s Algorithm and A* search were adapted to handle energy profiles.
Both approaches are based on a label-correcting strategy that incrementally links and merges energy profiles during the search. However, these methods often require complex rules to extend and relax energy profiles, adding significant overhead due to the inefficiency of explicitly merging profiles across the entire SoC range to maintain only one profile per graph vertex.
Continuing along this line, \citet{DBLP:journals/algorithmica/BaumDPSWZ20} extended their earlier work within a larger path planning context with charging stations, but still relied on the conventional label-correcting approach to handle uncertainty in initial SoC.
Despite recent progress in accelerating energy-optimal pathfinding queries with (preprocessing-based) speed-up techniques, the core problem of energy profile search remains largely unaddressed: existing profile search algorithms are still complex, often requiring specialized procedures and implementation tricks to enable the propagation of profiles during the search. This hinders their practical applicability and adoption.

\textbf{Contributions:} This paper presents a novel yet simple label-setting approach for energy-optimal profile search for EVs. The method adopts a multi-objective search strategy to propagate profiles in a best-first manner while incorporating efficient pruning mechanisms to discard unpromising profiles and to avoid expensive profile merging often observed in label-correcting schemes.
Experiment results on realistic instances demonstrate that our proposed approach offers a fast and practical framework for addressing energy-optimal profile queries in large road networks.

\section{Problem Definition and Background}
\label{sec:problem}
Consider a road network represented as a directed graph $G=(S,E)$, where $S$ is a finite set of states (intersection) and $E \subseteq S \times S$ is a set of edges (road segment).
Each edge has a real-valued energy cost, denoting the amount of energy required to traverse the link. Edge cost may be negative due to energy recuperation, and can be retrieved via the function $\mathit{cost}: E \rightarrow \mathbb{R}$.
A path $\pi$ in $G$ is defined as an ordered sequence of $n$ states $\{u_1,\dots,u_n\}$ where $(u_i,u_{i+1}) \in E$ for $ i \in \{1, \dots, n-1\}$.
Now, consider an EV starting with an initial energy of $\mathcal{E}_{init}\in\mathbb{R}^+$ at an initial location $\mathit{start} \in S$, and 
a maximum energy limit of $\mathcal{E}_{max}\in\mathbb{R}^+$ (battery capacity). 
The objective of energy-optimal pathfinding is to determine an optimum path $\pi^*$ that allows the EV to travel from $\mathit{start}$ to a target location $\mathit{goal} \in S$ with the least amount of energy consumed, while ensuring that the energy level of the EV remains within the $(0,\mathcal{E}_{max}]$ range at all locations along $\pi^*$.
If $\mathcal{E}_{\text{init}}$ is unknown, the objective becomes identifying, for each possible value of $\mathcal{E}_{\text{init}}$ within the allowable range, the corresponding energy-optimal path, i.e., computing all minimum-energy paths as a function of initial energy $\mathcal{E}_{\text{init}}$.

\begin{algorithm}[t]
\footnotesize
\caption{Energy-optimal A* Search}
\label{alg:Astar}
\DontPrintSemicolon
 \KwInput{A problem instance ($G$, $\mathit{cost}$, $h$, $\mathit{start}$, $\mathit{goal}$, $\mathcal{E}_{\mathit{init}}$)}
\KwOutput{Energy requirement of the optimal path}

$\mathit{Open} \gets \emptyset$, $\mathcal{C}(u) \gets \infty $ for all $u \in S$\;
$x \gets $ new node with $s(x) = \mathit{start}$ \;
${g}(x) \gets 0$, $f(x) \gets h(\mathit{start})$, $\mathcal{C}(\mathit{start}) \gets 0$, $ \mathcal{P}(x) \gets \varnothing$ \;
Add $x$ to $\mathit{Open}$\;
\While{$\mathit{Open} \neq \emptyset$}
{
Extract from $\mathit{Open}$ node $x$ with the smallest $f$-value \label{alg2:iter}\;
\lIf{$g(x) > \mathcal{C}(s(x)) $}
{ {\bf continue}}
     \lIf{$s(x)=\mathit{goal}$}
     {{\bf break} \label{alg2:terminate}  }
        \ForEach{$ v \in Succ(s(x))$}
        {
        $y \gets $ new node with $s(y) = v$ \;       
        ${g}(y) \gets {g}(x) + \mathit{cost}(s(x),v)$ \;
        $f(y) \gets g(y) + h(v)$ \;
        $\mathcal{P}(y) \gets x$ \;
        \lIf{$g(y) > \mathcal{E}_{init}$ {\bf or} $f(y) > \mathcal{E}_{init}$ \label{astar:energy-check2}}
            { {\bf continue}}
        \If{$ \mathcal{E}_{init} - g(y) > \mathcal{E}_{max}$ \label{astar:energy-check1}}
            {$g(y) \gets \mathcal{E}_{init} - \mathcal{E}_{max}$}
        \If{$g(y) < \mathcal{C}(v) $}
            { 
            $\mathcal{C}(v) \gets g(y)$ \;
            Add $y$ to $Open$ \label{astar:relax-check2}\;
            }
        }
}
\Return{$\mathcal{C}(\mathit{goal})$}\;
\end{algorithm}
\subsection{Energy-optimal pathfinding with A*}
We now briefly review energy-optimal pathfinding on the basis of A*. Following the heuristic search literature, we define partial paths to states as search nodes. More precisely, a node $x$ represents a path originating from $\mathit{start}$.
Each node $x$ is associated with a state $s(x)$ (i.e., the last state in the partial path), a cost value $g(x)$ representing the cumulative energy consumed from $\mathit{start}$ to $s(x)$, a reference $\mathcal{P}(x)$ pointing to its parent node, and a total cost estimate $f(x)$, a lower bound on energy cost from $\mathit{start}$ to $\mathit{goal}$ via $x$.

The search in A* is guided by nodes with the smallest cost estimates, or $f$-values, which are conventionally computed using a consistent and admissible heuristic function $h: S \rightarrow \mathbb{R}$ such that $f(x) = g(x) + h(s(x))$~\cite{hart1968formal}. $h$ is consistent if $h(u) \le h(v) + \mathit{cost}(u,v)$ for every edge $(u,v) \in E$, and also admissible if we additionally have $h(\mathit{goal}) = 0$.
Consistent heuristics offer a key advantage in negative-weight graphs: $f$-values in A* are still guaranteed to increase monotonically during the search~\cite{AhmadiSHJ24}. This property inherently eliminates the need for reweighting techniques and provides a robust framework for solving the problem in dynamic settings, where both energy models and heuristic functions may change between queries~\cite{AhmadiRJ25_ESA25}. However, this benefit comes at an additional computational cost, as some better informed heuristic functions may involve non-linear and potentially expensive computations.

As discussed in~\cite{SchonfelderLW14}, consistent energy heuristic functions for EVs can be naturally derived using physical energy models, typically formulated as lower bounds on the sum of kinetic and potential energy. In this context, the kinetic energy estimate is modeled as a function of air-line (Euclidean) distance, while the potential energy component is based on elevation differences.
We have provided two such heuristic functions in the Appendix, derived from model-based and data-driven energy consumption models.

Algorithm~\ref{alg:Astar} presents a pseudocode for energy-optimal A* search, adapted from~\citet{SchonfelderLW14}.
It begins by initializing a node with the $\mathit{start}$ state and inserting it into $\mathit{Open}$, a priority queue that stores all generated but unexplored nodes. The algorithm also initializes a scalar $\mathcal{C}(u)$ for every $u \in S$ to track states' best-known energy cost during the search.
It then proceeds by expanding nodes in order of $f$-values. Expanding a node $x$ involves extending its partial path to each of its successor states, denoted by $\mathit{Succ}(x)$, thereby generating descendant nodes. 
During expansion, each descendant node $y$ is checked for feasibility and, if necessary, cost adjustment to ensure the available energy at each state remains within the EV's battery capacity constraints \cite{artmeier2010shortest}. 
We can distinguish two special cases: i) if the energy requirement $g(y)$ or estimated total cost $f(y)$ of the descendant node exceeds the available initial energy $\mathcal{E}_{\mathit{init}}$, $y$ is considered infeasible and can be discarded; ii) if the remaining energy at the successor state, computed as $\mathcal{E}_{\mathit{init}} - g(y)$, exceeds the EV’s maximum battery capacity (e.g., on sharp downhill segments), then $g(y)$ is adjusted to cap the recuperated energy. This ensures that the remaining energy does not exceed $\mathcal{E}_{\mathit{max}}$, meaning that some recuperated energy cannot be stored due to battery limitations.
If a lower-cost path to $s(y)$ is found, the value $\mathcal{C}(s(y))$ is updated, and $y$ is inserted into $\mathit{Open}$ for further exploration. The search terminates when a node with the $\mathit{goal}$ state is extracted from $\mathit{Open}$, with $\mathcal{C}(\mathit{goal})$ denoting the minimum energy cost. If $\mathcal{C}(\mathit{goal}) = \infty$, it indicates that no feasible path exists under the given initial energy $\mathcal{E}_{\mathit{init}}$.

\subsection{Pathfinding with unknown $\mathcal{E}_{init}$}
A common method to address uncertainty in the initial energy level is to define the energy requirement of an edge $(u,v)$ as a function that maps any possible initial energy $\mathcal{E}_{\mathit{init}} \in [0, \mathcal{E}_{\mathit{max}}]$ at node $u$ to its corresponding energy cost. The energy profile of a path can then be derived by composing the energy profiles of its constituent edges \cite{baum2013energy}.
As shown in \citet{eisner2011optimal} and formally proven in \citet{DBLP:journals/algorithmica/BaumDPSWZ20}, the energy profiles of both edges and paths can be represented as piecewise-linear functions with up to two breakpoints and fixed slopes. In our notation, we define the breakpoints of a path profile represented by node $x$ as $(\mathcal{E}_{\mathit{min}}(x), g(x))$ and $(\mathcal{E}_{\mathit{max}}, \overline{g}(x))$, where $\mathcal{E}_{\mathit{min}}(x)$ denotes the minimum energy required to traverse the path, and $g(x)$ and $\overline{g}(x)$ represent the minimum and maximum energy costs, respectively. Here, $\overline{g}(x)$ corresponds to the energy cost when all or part of the recuperated energy cannot be stored due to the capacity limit $\mathcal{E}_{\mathit{max}}$.

\begin{figure}[t]
\footnotesize
\centering 
\includegraphics[width=1\columnwidth]{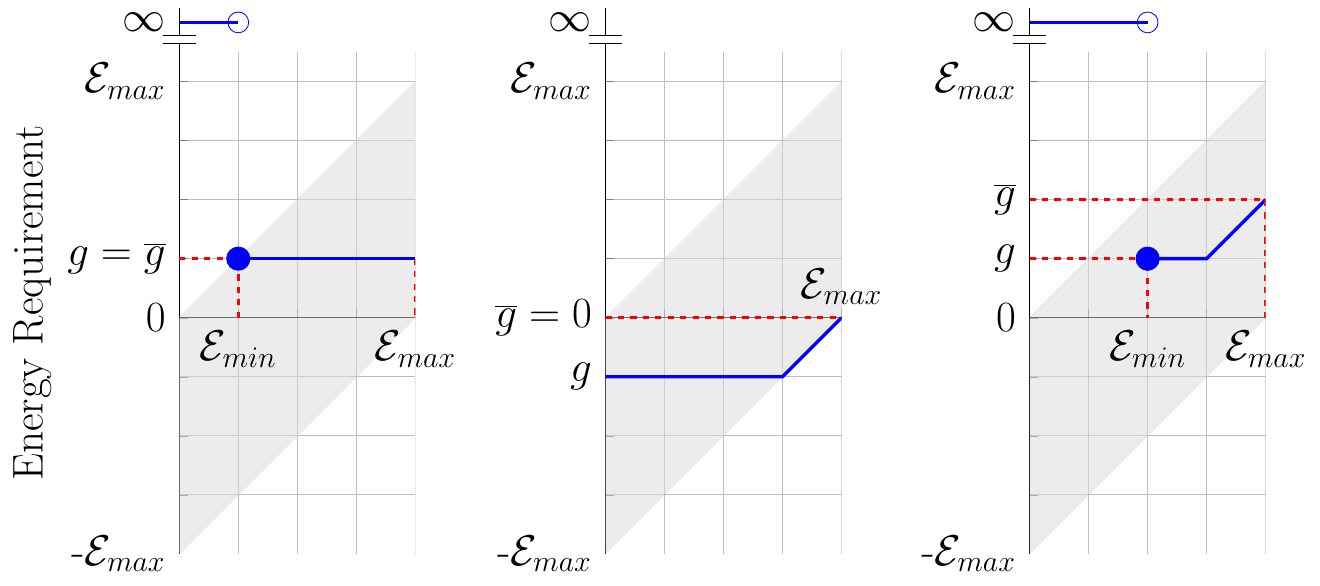} 
\caption{Schematic illustration of energy profiles. Horizontal axis is the $\mathcal{E}_{\mathit{init}}$ range. From left to right: positive-cost link, negative-cost link, and a generic path profile.}
\label{path_profiles_def}
\end{figure}

Figure~\ref{path_profiles_def} illustrates a generic path profile, including examples of negative-cost links, positive-cost links, and generic paths. The shaded area denotes the feasible range. A generic profile consists of two distinct segments:\\
(i)~$\mathcal{E}_{\mathit{init}} < \mathcal{E}_{\mathit{min}}$: the path is not traversable, and the energy cost is considered infinite;\\
(ii)~$\mathcal{E}_{\mathit{init}} \geq \mathcal{E}_{\mathit{min}}$: the path is traversable with a minimum energy cost of $g(x)$ (first segment), but may increase linearly up to $\overline{g}(x)$ as $\mathcal{E}_{\mathit{init}}$ increases (second segment).
In the case of negative-energy links, for example, the path can be traversed even with $\mathcal{E}_{\mathit{init}} = 0$. However, if $\mathcal{E}_{\mathit{init}} = \mathcal{E}_{\mathit{max}}$, no energy can be recuperated, resulting in a net energy cost of zero. An important property of these energy profiles is that the slope of the second segment is always one. In other words, at higher initial energy levels, the energy cost increases exactly by the amount of energy that the EV is unable to store~\cite{storandt2012quick}.
Also note that the following inequalities always hold: $\overline{g}(x) \geq g(x)$ and $\mathcal{E}_{\mathit{min}}(x) \geq g(x)$, since the EV must have at least $g(x)$ units of energy to be able to cover the path’s minimum cost. In simple cases, such as paths without negative-cost edges (e.g., left plot of Figure~\ref{path_profiles_def}), we may observe flat profiles with $\mathcal{E}_{\mathit{min}}(x) = g(x) = \overline{g}(x)$.

The profile \textit{linking} procedure involves shifting energy profiles in both the \textit{cost} and $\mathcal{E}_{\mathit{init}}$ domains by the energy cost of the added edge, while respecting the battery capacity constraints~\cite{eisner2011optimal, baum2013energy, SchonfelderLW14}. Linking a path profile with an edge profile is a straightforward task and can be performed in constant time as we only need to determine new breakpoints after the profiles are linked, given that the resulting path profile follows the generic piecewise-linear form above.
However, the specific calculations vary slightly depending on the direction of the search, whether a forward or backward search is being performed. In the next section, we describe the linking operation for a forward search. 
An illustrative example of profile linking in the backward direction is provided in the Appendix for improved clarity.

\section{Energy-Optimal Profile {A*} Search}
Existing label-correcting profile search methods employ a specialized procedure, called \textit{merge}, to combine multiple energy profiles associated with the same state. In practice, merging two profiles involves scanning the entire range of initial energy level and computing a new profile that represents the lower envelope of the input profiles, ensuring the minimum possible cost is selected for every value of $\mathcal{E}_{\mathit{init}}$, sometimes by discretizing the SoC range~\cite{eisner2011optimal}.
If the merged profile introduces different breakpoints compared to the existing one, the corresponding state will be re-expanded with the new profile. 
Although each individual path profile can be described using at most two breakpoints, merged profiles may exhibit up to $|S|$ breakpoints ~\cite{DBLP:journals/algorithmica/BaumDPSWZ20}.
Despite this bound, the \textit{merge} operation remains a complex and non-trivial task, requiring careful implementation to ensure efficiency.

We now introduce our novel A*-based approach, which leverages efficient pruning rules to streamline profile search. While the method can be applied in both forward and backward directions, we present it here in the conventional forward search formulation.
Inspired by recent advances in multi-objective heuristic search~\cite{LTMOA3,ahmadi_ICAPS25}, in particular scenarios with negative weights~\cite{AhmadiSHJ24,AhmadiRJ25_ESA25}, we frame energy-optimal profile search as a multi-objective shortest path problem with negative costs \cite{StewartW91}, where the minimum required energy ($\mathcal{E}_{\mathit{min}}$) and the maximum energy cost ($\overline{g}$) act as additional objectives alongside the primary objective of minimizing energy cost ($g$). This formulation enables us to efficiently identify and prune unpromising paths during best-first search, effectively bypassing the need for the costly \textit{merge} operation.
Before describing the main steps of the algorithm, we introduce the key concept of \textit{dominance} between search nodes.

\noindent \textbf{Definition.\ }
A node $y$ is said to be \textit{dominated} by node $x$ if $\mathcal{E}_{\mathit{min}}(x) \leq \mathcal{E}_{\mathit{min}}(y)$, ${g}(x) \leq {g}(y)$, and $\overline{g}(x) \leq \overline{g}(y)$.

\begin{figure*}[t]
\footnotesize
\centering 
 \includegraphics[width=1\textwidth]{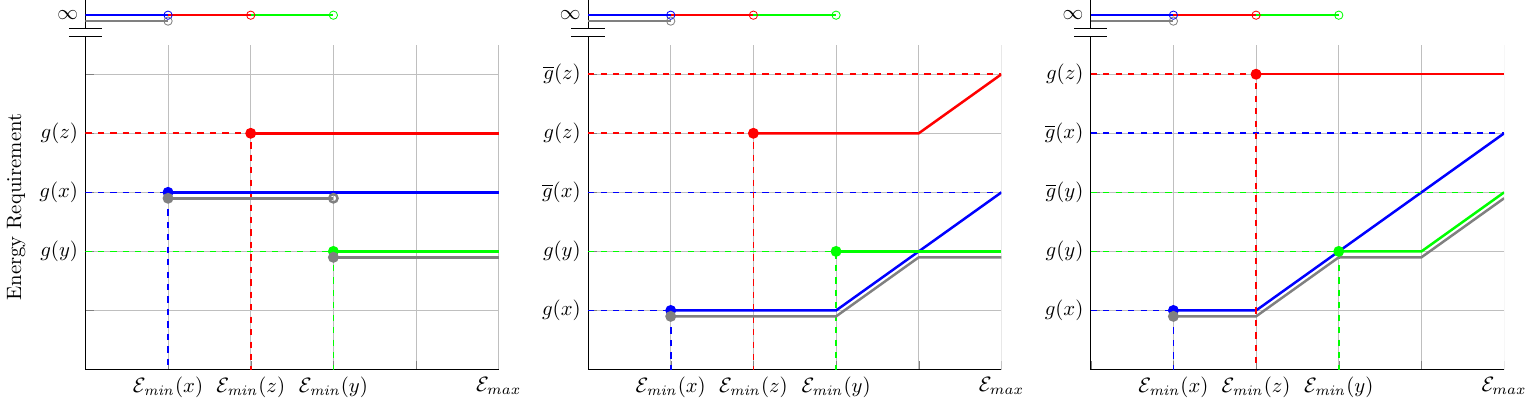} \\
\caption{Dominance and non-dominance of energy profiles. In each plot, the profile of $z$ (in red) is dominated by the profile of $x$ (in blue). The profiles of $x$ and $y$ (blue and green) are non-dominated and can both contribute to the lower envelope (in gray).}
\label{nondominated_profiles}
\end{figure*}

\noindent
Intuitively, this means that the energy profile of node $y$ lies entirely above or equal to that of node $x$ across all relevant initial energy levels, and thus $y$ does not offer a lower energy cost at any point in the $(0, \mathcal{E}_{\mathit{max}}]$ range. Figure~\ref{nondominated_profiles} illustrates three different scenarios involving energy profiles associated with nodes $x$, $y$, and $z$, where the profile of node $z$ (shown in red) is dominated by that of $x$. For instance, in the middle plot, node $z$ is dominated by node $x$ since there exists no initial energy level at which $z$ yields a lower energy cost.
In contrast, nodes $x$ and $y$ are non-dominated, as neither profile entirely dominates the other: each offers a lower cost over some portion of the initial energy range (vertical axis), and therefore both may contribute to the lower envelope.

\begin{algorithm}[t]
\small
\caption{Energy Profile (Forward) A* Search}
\label{alg:Astarfw}
\DontPrintSemicolon
 \KwInput{A problem instance ($G$, $\mathit{cost}$ , $h$, $\mathit{start}$, $\mathit{goal}$)}
 \KwOutput{A superset of nodes representing optimal profiles}
 $\mathit{Open} \gets \emptyset$, \ $\overline{f} \gets \infty$ \;
$\mathcal{X}(u) \gets \emptyset$ $\forall u \in S$\;
 $x \gets $ new node with $s(x) = \mathit{start}$\ \;
 $g(x) \gets 0 $, $\overline{g}(x) \gets 0$, $\mathcal{E}_{\mathit{min}}(x) \gets 0$, $ parent(x) \gets \varnothing$ \;
 $f(x) \gets h(\mathit{start}) $\;
Add $x$ to $Open$\;
\While{$Open \neq \emptyset$\label{alg:Astarfw:iter}}
{
 Extract from $Open$ a node $x$ with the smallest $f$-value \;
     \lIf{ $f(x) \geq \overline{f}$}
     { {\bf break} }
     
       $\mathit{dominated} \gets \mathrm{false}$\label{alg:Astarfw:dom1}\;
     \ForEach{$y \in \mathcal{X}(s(x))$}
        { 
       \If{$\mathcal{E}_{\mathit{min}}(y) \leq \mathcal{E}_{\mathit{min}}(x) $ {\bf and} $\overline{g}(y) \leq \overline{g}(x) $}
            {
            $\mathit{dominated} \gets \mathrm{true}$ \; 
        {\bf break}\label{alg:Astarfw:removedom2}} 
        }
     \lIf{$\mathit{dominated} = \mathrm{true}$} {\textbf{continue}\label{alg:Astarfw:dom2}}

     add $x$ to $\mathcal{X}(s(x))$\label{alg:Astarfw:add}\;
     
     \If{$s(x)=\mathit{goal}$\label{alg:Astarfw:goal1}}
     { 
     
     $\overline{f} \gets \min (\overline{f},\max(\mathcal{E}_{\mathit{min}}(x), \overline{g}(x)))$\label{alg:Astarfw:goal3}\;
     {\bf continue}  \;\label{alg:Astarfw:goal2}}

    \ForEach{$v \in Succ(s(x))$\label{alg:Astarfw:exp1}}
        { 
        $y \gets $ new node with $s(y) = v$ \;       
        $cost_{e} \gets \mathit{cost}(s(x),v)$ \;
        ${g}(y) \gets \max({g}(x) + \mathit{cost}_{e},  -\mathcal{E}_{max})$ \;
        $f(y) \gets g(y) + h(v)$ \;
        $\mathcal{E}_{\mathit{min}}(y) \gets \max(\mathcal{E}_{\mathit{min}}(x), {g}(y)$) \;
        $\overline{g}(y) \gets \max(0,\overline{g}(x)+cost_{e})$ \;
        $parent(y) \gets x$ \;
        \If{$\mathcal{E}_{\mathit{min}}(y) > \mathcal{E}_{max}$ {\bf or} $\overline{g}(y) > \mathcal{E}_{max}$ \label{alg:Astarfw:energy-check1}}
            { {\bf continue}}
        \lIf{$\overline{g}(y) + h(v) > \mathcal{E}_{max}$ \label{alg:Astarfw:energy-check2}}
            { {\bf continue}}
        $z \gets $ last node in $\mathcal{X}(v)$ \label{alg:Astarfw:lazy1}\;
        \If{$\mathcal{E}_{\mathit{min}}(z) \leq \mathcal{E}_{\mathit{min}}(y) $ {\bf and} $\overline{g}(z) \leq \overline{g}(y) $}
            { {\bf continue}\label{alg:Astarfw:lazy2}}
        
        Add $y$ to $Open$ \label{alg:Astarfw:add2}\;
        
        }
       
}
\Return{$\mathcal{X}(\mathit{goal})$}\; 
\end{algorithm}

A pseudocode of our A*-based profile search is presented in Algorithm~\ref{alg:Astarfw}.
It starts with initializing a priority queue $\mathit{Open}$, which traditionally stores unexplored (\textit{Open}) nodes of the search, and a cost upper bound $\overline{f}$.
It then sets up for every state $u\in S$ a list $\mathcal{X}(u)$, which is responsible for storing all non-dominated nodes expanded with $u$. 
Since the search is done in the forward direction, the algorithm generates an initial (zero cost) node associated with $\mathit{start}$ and inserts it into $\mathit{Open}$.
The main search starts at line~\ref{alg:Astarfw:iter}.
In every iteration, A* extracts from $\mathit{Open}$ a node with the smallest $f$-value among all nodes present in the queue.
Let this extracted node be $x$. 
We have $f(x) = g(x) + h(x)$, where $g(x)$ represents the minimum cost of the concrete path from $\mathit{start}$ to $s(x)$. 
We now discuss our main pruning rule.

\noindent
\textbf{Dominance pruning rule:}
Let $x$ and $y$ be two nodes associated with the same state, and without loss of generality, assume that $x$ is extracted after $y$ has been expanded. This implies $f(y) \leq f(x)$ due to A*  processing nodes in non-decreasing order of $f$-values. Given that the primary cost is already ordered by A* (one dimension reduced), node $x$ is dominated by node $y$ if $\overline{g}(y) \leq \overline{g}(x)$ and $\mathcal{E}_{\mathit{min}}(y) \leq \mathcal{E}_{\mathit{min}}(x)$, meaning that $x$ does not offer a better energy cost than $y$ at any $\mathcal{E}_{\mathit{init}}$ within the range.
Now let $\mathcal{X}(s(x))$ denote the set of previously expanded (non-dominated) nodes associated with state $s(x)$. Then, each newly extracted node $x$ can be rigorously checked against nodes in $\mathcal{X}(s(x))$ to determine whether $x$ should be pruned (lines~\ref{alg:Astarfw:dom1}-\ref{alg:Astarfw:dom2}), avoiding expansions that are guaranteed not to contribute to any energy-optimal path.
However, there is an important observation in this form of label pruning that distinguishes it from label-correcting approaches: it is possible for a node $x$ to be non-dominated with respect to any individual node in $\mathcal{X}(s(x))$, but still be dominated by the lower envelope formed by the set as a whole. This means that $\mathcal{X}(s(x))$ may contain nodes that do not themselves contribute to the true lower envelope. 
One might consider a more aggressive pruning strategy by explicitly maintaining the true lower envelope of profiles (via profile merging), as done in label-correcting approaches. However, as we demonstrate in the experimental section, our relaxed pruning rule, though less rigorous, achieves comparable performance to conventional (non-profile) A* search while avoiding the overhead of constructing and maintaining per-state lower envelopes.

Node $x$ is added to the $\mathcal{X}(s(x))$ set if it is deemed non-dominated (line~\ref{alg:Astarfw:add}). If $x$ is associated with the $\mathit{goal}$ state, then a tentative optimal path has been found  (lines~\ref{alg:Astarfw:goal1}-\ref{alg:Astarfw:goal2}), and we can attempt to improve the search upper bound $\overline{f}$ on the solution cost  (line~\ref{alg:Astarfw:goal3}).
Let $x$ be such a tentative solution node. Then, any future node $y$ with estimated cost $f(y)$ no smaller than either $\mathcal{E}_{\mathit{min}}(x)$ or $\overline{g}(x)$ can be safely skipped. This pruning is valid because we would have $f(x) \leq \mathcal{E}_{\mathit{min}}(x) \leq f(y)$ and $f(x) \leq \overline{g}(x) \leq f(y)$ for any such out-of-bounds node $y$ (note that $f(x) = g(x)$ for solution paths). Furthermore, since $f(y) = g(y) + h(s(y)) \leq \overline{g}(y) + h(s(y))$ and likewise $f(y) \leq \mathcal{E}_{\mathit{min}}(y) + h(s(y))$, any solution path obtained by extending $y$ is guaranteed to be dominated by the existing solution $x$.
Thus, the maximum of $\mathcal{E}_{\mathit{min}}(x)$ and $\overline{g}(x)$ can be treated as an upper bound, which the algorithm can dynamically update when a tighter bound is discovered.

If $x$ is not a solution node, it will be expanded through its successor states (lines~\ref{alg:Astarfw:exp1}-\ref{alg:Astarfw:add2}). Let $v$ be one such successor state, i.e., $(s(x), v) \in E$, with its energy requirement denoted by $\mathit{cost}_e$. Each descendant node generated during this expansion corresponds to a new energy profile, obtained by vertically shifting the profile of $x$ by $\mathit{cost}_e$ and adjusting the resulting breakpoints to ensure they remain within the battery constraints.
As in energy A* search, we compute for each descendant node $y$ its (minimum) energy cost $g(y)$, bounded by $-\mathcal{E}_{max}$ to avoid recuperation more than the battery capacity, and estimated energy cost $f(y)$. 
We then compute $\mathcal{E}_{\mathit{min}}(y)$, the minimum initial energy required (at $\mathit{start}$) to traverse the extended path. 
This is calculated as the maximum of $g(y)$ and the minimum energy required to traverse the prefix path represented by $x$, i.e., $\mathcal{E}_{\mathit{min}}(x)$. 
In other words, $\mathcal{E}_{\mathit{min}}(y)$ represents the maximum non-negative cumulative energy cost observed over any prefix of $y$.
This greedy strategy ensures that the minimum required energy does not decrease along the path, even when encountering negative-cost links, since the energy needed to traverse the earlier subpath still applies. Consequently, we always have $g(y) \leq \mathcal{E}_{\mathit{min}}(y)$ in our forward profile search.

To compute $\overline{g}(y)$, we shift $\overline{g}(x)$ by adding $\mathit{cost}_e$, while ensuring that the result remains non-negative to avoid the infeasible scenario of starting a path with a fully charged battery and ending with more than 100\% SoC. 
More importantly, this lower-bounding of $\overline{g}(y)$ enables us to greedily capture the maximum energy cost observed among all subpaths of $y$ ending at $v$, i.e., the suffix of $y$ with the largest (non-negative) cost.
This forms a natural pruning rule: if $\overline{g}(y) > \mathcal{E}_{\mathit{max}}$, then there exists a subpath within $y$ that cannot be traversed even with a fully charged battery, implying that $y$ is infeasible.
Note that the remaining energy at $v$ when starting with $\mathcal{E}_{\mathit{init}} = \mathcal{E}_{\mathit{max}}$ is given by $\mathcal{E}_{\mathit{max}} - \overline{g}(y)$.

With the energy profile of the descendant node $y$ computed, we then perform feasibility checks to determine whether $y$ can lead to a valid solution (lines~\ref{alg:Astarfw:energy-check1},~\ref{alg:Astarfw:energy-check2}). Two simple conditions are checked: 
(i) Node $y$ is considered infeasible if its $\mathcal{E}_{\mathit{min}}$ or $\overline{g}$ exceeds the battery capacity $\mathcal{E}_{\mathit{max}}$, meaning the extended path to state $v$ cannot be traversed even with a fully charged battery; 
(ii) Node $y$ is also pruned if its estimated total energy cost (i.e., maximum suffix cost of $y$ plus the cost-to-go from $v$ to $\mathit{goal}$) exceeds $\mathcal{E}_{\mathit{max}}$.
These checks ensure that only nodes satisfying the battery constraints are further explored.

Descendant nodes can be added to $\mathit{Open}$ if their energy profiles fall within the battery constraints, and if they are not dominated by any previous expansions with state $v$. However, instead of rigorously verifying dominance during expansion, our approach delays this check and performs it \textit{lazily} when nodes are extracted from the queue.
In addition, to avoid overloading the queue with dominated labels, we apply a quick, constant-time dominance check during expansion by comparing the descendant node $y$ with the most recently expanded node associated with the same successor state $v$ (lines~\ref{alg:Astarfw:lazy1}-\ref{alg:Astarfw:lazy2}). This pruning helps reduce unnecessary insertions while preserving correctness.
A similar lazy strategy has been used in the context of multi-objective search~\cite{LTMOA3,AhmadiSHJ24}.

Finally, the algorithm terminates either by surpassing the upper bound $\overline{f}$ or when $\mathit{Open}$ becomes empty, and returns $\mathcal{X}(\mathit{goal})$: a superset of nodes associated with energy-optimal paths between $\mathit{start}$ and $\mathit{goal}$. If $\mathcal{X}(\mathit{goal})$ is empty upon termination, no feasible solution exists for the given problem instance.
To choose form optimal paths, one can compute the minimum energy cost for any given initial energy level $\mathcal{E}_{\mathit{init}}$ through the energy profile of nodes in $\mathcal{X}(\mathit{goal})$. 
The optimum cost can then be obtained either by selecting a profile that yields the lowest cost at $\mathcal{E}_{\mathit{init}}$, or by computing the lower envelope of all profiles in $\mathcal{X}(\mathit{goal})$ as a one-time operation.
Since all nodes retain full backtracking information, reconstructing solution paths is as simple as following the parent pointers from $\mathit{goal}$ back to $\mathit{start}$.

\subsection{Theoretical Results}
This section provides a formal proof of correctness for our energy profile search algorithm, which we call Pr-A*.

\vspace{0.5em}
\noindent \textbf{Lemma 1.}
Energy profile of every path can be sufficiently represented by two breakpoints: $(\mathcal{E}_{\mathit{min}}, g)$ and $(\mathcal{E}_{\mathit{max}}, \overline{g})$.

\noindent \textbf{Proof Sketch.}
\citet{eisner2011optimal} showed that the number of breakpoints required to represent energy profiles is bounded. Furthermore, Case 3 of Lemma~1 in~\citet{DBLP:journals/algorithmica/BaumDPSWZ20} proves that such profiles are piecewise linear with fixed slopes and can be fully represented using just two breakpoints (we handle their Cases 1,2 within the algorithm).
While our formulation slightly differs in terms of profile definition from~\citet{DBLP:journals/algorithmica/BaumDPSWZ20}, which maps initial to final energy level, the conceptual basis remains equivalent, as the energy cost in our formulation is simply the difference between the final and initial energy levels.

In our notation, the first breakpoint, $\mathcal{E}_{\mathit{min}}$, denotes the minimum initial energy required to traverse the path at its minimum energy cost $g$. The second breakpoint is fixed at $\mathcal{E}_{\mathit{max}}$ and corresponds to the maximum energy cost $\overline{g}$. These two points are sufficient to define the entire feasible range of energy profiles, which can be reconstructed using two linear segments: one with zero slope (constant cost $g$) and the second with unit slope (increasing cost to $\overline{g}$). The transition point between these segments is at $(\mathcal{E}_{\mathit{max}} - \overline{g} + g, g)$, which can be obtained from the slope structure of the profile. 
\hfill $\square$

\vspace{0.5em}
\noindent \textbf{Lemma 2.}
Suppose Pr-A* expands nodes in non-decreasing order of $f$-values (which may be negative). Let $x_i$ and $x_{i+1}$ be two nodes extracted from $\mathit{Open}$ in consecutive iterations. Then, $f(x_i) \leq f(x_{i+1})$ if $h$ is consistent.

\noindent \textbf{Proof Sketch.}
There can be two cases:
(i) If $x_{i+1}$ was already in $\mathit{Open}$ when $x_i$ was extracted, the claim follows directly.  
(ii) Otherwise, if $x_{i+1}$ is a descendant of $x_i$, then by consistency we have:
\[
h(s(x_i)) \leq h(s(x_{i+1})) + \mathit{cost}(s(x_i), s(x_{i+1}))
\]
Adding $g(x_i)$ to both sides yields $f(x_i) \leq f(x_{i+1})$.
\hfill $\square$

\vspace{0.5em}
\noindent \textbf{Corollary 1.}
Let $(x_1, x_2, \dots, x_t)$ be the sequence of nodes extracted from $\mathit{Open}$. If the heuristic function $h$ is consistent and admissible, then $i \leq j$ implies $f(x_i) \leq f(x_j)$, i.e., the $f$-values of extracted nodes are monotonically non-decreasing \cite{hart1968formal}.

\vspace{0.5em}
\noindent \textbf{Lemma 3.}
Let nodes $x$ and $y$ be associated with the same state, and suppose $x$ is extracted before $y$. $x$ dominates $y$ if:
\[
\mathcal{E}_{\mathit{min}}(x) \leq \mathcal{E}_{\mathit{min}}(y) \quad \text{and} \quad \overline{g}(x) \leq \overline{g}(y)
\]

\noindent \textbf{Proof Sketch.}
From Corollary~1, we have $f(x) \leq f(y)$. Since $s(x) = s(y)$, we have $h(s(x)) = h(s(y))$, implying $g(x) \leq g(y)$. The dominance conditions on $\mathcal{E}_{\mathit{min}}$ and $\overline{g}$ confirm that $x$ offers a strictly better or equal profile at all energy levels, thus $y$ is dominated by $x$.
\hfill $\square$

\vspace{0.5em}
\noindent \textbf{Lemma 4.}
Let $y$ be a node dominated by a previously explored node $x$, both associated with the same state. Then the expansion of $y$ is not necessary.

\noindent \textbf{Proof Sketch.}
Assume, for contradiction, that expanding $y$ is necessary to obtain a cost-optimal or non-dominated solution. Since $x$ dominates $y$, we have:
\[
g(x) \leq g(y), \quad \mathcal{E}_{\mathit{min}}(x) \leq \mathcal{E}_{\mathit{min}}(y), \quad \overline{g}(x) \leq \overline{g}(y)
\]
This implies that any path extended from $y$ (including any solution path) can be replaced with an equal or better path extended from $x$, contradicting the necessity of expanding $y$ to obtain a cost-optimal solution path.
\hfill $\square$

\vspace{0.5em}
\noindent \textbf{Lemma 5.}
Node $y$ can be pruned if:
\[
\mathcal{E}_{\mathit{min}}(y) > \mathcal{E}_{\mathit{max}} \ \text{or} \
\overline{g}(y) > \mathcal{E}_{\mathit{max}} \ \text{or} \ \overline{g}(y) + h(s(y)) > \mathcal{E}_{\mathit{max}}
\]

\noindent \textbf{Proof Sketch.}
The first condition directly violates the battery constraint on maximum available initial energy. 
The second case indicates that the cost of traversing the path represented by $y$ is more than EV's maximum possible available energy even with a fully charged battery, violating the battery capacity. 
In the third case, since $h$ is consistent and admissible, we can guarantee that extending the maximum cost suffix of $y$ toward $\mathit{goal}$ would yield a subpath exceeding the energy capacity, rendering the expansion of $y$ unnecessary.

\vspace{0.5em}
\noindent \textbf{Lemma 6.}
Let $x$ be a tentative solution node. Then, expanding any node $y$ with $f(y) > \max\{\mathcal{E}_{\mathit{min}}(x), \overline{g}(x)\}$ is not necessary.

\noindent \textbf{Proof Sketch.}
For such a node $y$, we have:

$f(x) \leq \mathcal{E}_{\mathit{min}}(x) \leq f(y)$ and $f(x) \leq \overline{g}(x) \leq f(y)$

Given that $f(y) = g(y) + h(s(y))$, we have:
\[
f(y) \leq \overline{g}(y) + h(s(y)) \\ \ \text{and} \\ \ f(y) \leq \mathcal{E}_{\mathit{min}}(y) + h(s(y)),
\]
It follows that any solution path extended from $y$ will be dominated by the existing solution node $x$. This is because:
\[
\mathcal{E}_{\mathit{min}}(x) \leq \mathcal{E}_{\mathit{min}}(y) + h(s(y)) \\ \ \text{and} \\ \ g(x) \leq \overline{g}(y) + h(s(y))
\]
Thus, if $f(y)$ exceeds both $\mathcal{E}_{\mathit{min}}(x)$ and $\overline{g}(x)$, any further expansion from $y$ cannot produce a better or non-dominated solution profile. Therefore, $y$ can be safely pruned.
\hfill $\square$

\vspace{0.5em}
\noindent \textbf{Theorem 1.}
Let $h$ be a consistent and admissible energy heuristic function. Then Pr-A* returns a superset of optimal nodes representing the energy-optimal profiles from $\mathit{start}$ to $\mathit{goal}$, if a feasible path exists.

\noindent \textbf{Proof Sketch.}
The algorithm explores all feasible nodes toward $\mathit{goal}$ in best-first order. Dominated nodes are identified using Lemma~3 and safely pruned according to Lemma~4. Nodes with infeasible or out-of-bounds energy estimates are pruned based on Lemma~5. Furthermore, the global upper bound $\overline{f}$ provides a safe early termination criterion, since all nodes remaining in $\mathit{Open}$ can be guaranteed not to yield an optimal solution (Lemma~6).
Upon termination, the resulting set $\mathcal{X}(\mathit{goal})$ contains all individually non-dominated solution nodes. While some nodes in this set may appear dominated when compared to the global lower envelope of optimal profiles, the set nonetheless forms a superset of all truly optimal nodes.
Consequently, all energy-optimal paths across the full range of $\mathcal{E}_{\mathit{init}}$ are guaranteed to be represented. The final lower envelope of optimal profiles can be efficiently extracted via post-processing over $\mathcal{X}(\mathit{goal})$.
\hfill $\square$

\section{Experimental Analysis}
This section evaluates the performance of our energy-optimal profile search framework on real-world road networks.
We used 10 large maps from the 9\textsuperscript{th} DIMACS Implementation Challenge\footnote{\url{http://www.diag.uniroma1.it/challenge9}} as benchmark graphs, with the largest map (Central USA) containing over 14 million states and 34 million edges. The map data includes distance and travel time per edge. We enriched these maps with elevation data from the Shuttle Radar Topography Mission\footnote{\url{https://www2.jpl.nasa.gov/srtm/}}.
For each link in the network, we compute energy consumption using the energy model presented in~\citet{ahmadi2024realtime}, which accounts for road gradient, vehicle mass, and speed profiles. We simulate a Nissan Leaf carrying four passengers and equipped with an extended-range 85\,kWh battery. All graphs are negative-cycle free.
To evaluate performance, we generate 200 random queries per map, yielding a total of 2000 energy-optimal pathfinding queries for our experiments.

\textbf{Implementation:}
We implemented four variants of our energy profile search algorithm in C++, alongside standard energy-optimal Dijkstra and A* search algorithms, which work with a known initial SoC. Our four variants include:

\begin{itemize}
    \item \textbf{Pr-A*\textsubscript{fw}:} A forward energy profile search based on A*, as described in Algorithm~\ref{alg:Astarfw}.

    \item \textbf{Pr-A*\textsubscript{bw}:} A backward energy profile search using A*, which explores the graph from $\mathit{goal}$ to $\mathit{start}$ using predecessors. This variant follows a similar expansion and pruning procedure as the forward version, but differs slightly in how profiles are linked and checked. 
    A pseudocode of this variant is provided in the Appendix.

    \item \textbf{Pr-BA*:} A bidirectional profile search using A*, employing an interleaved strategy that incorporates both forward and backward searches above. Complete paths are constructed by joining forward and backward nodes that meet at the same state. In each iteration, the algorithm selects a node from either direction based on the smallest estimated cost. Nodes can be joined with all explored nodes from the opposite direction (that reach the same state) to form complete paths and potentially update the upper bound if feasible. A pseudocode of this variant is provided in the Appendix.

    \item \textbf{Pr-BA*\textsubscript{par}:} A parallel version of Pr-BA*, where each search direction is executed on a separate CPU thread. Each thread performs best-first exploration independently, applying the same pruning and joining rules as in the non-parallel variant. 
\end{itemize}
All variants were implemented using the same data structures, and binary heap as the priority queue, along with identical heuristic functions to enable a fair, head-to-head comparison. 
For the Dijkstra implementation, we equipped it with a potential energy function (identical to the one used in our heuristic function) to enable online reweighting of edge costs without preprocessing, as discussed in~\citet{ahmadi2024realtime}.
For Dijkstra and A* searches, we assume initial SoC of 100\%.
All C++ code was compiled using the \texttt{GCC 11.04} compiler with optimization level \texttt{-O3}. Experiments were conducted on a machine equipped with an Intel Xeon Platinum 8488C processor @3.2\,GHz and 16\,GB of RAM. Each instance was run three times, and we report the results corresponding to the run with the median runtime.
Our code is publicly available\footnote{https://bitbucket.org/s-ahmadi/eospp}.

\begin{table}[t]
\caption{Runtime statistics of the algorithms (in milliseconds) over all feasible instances, separated by four energy ranges. The table also reports speedup ($\eta$) relative to Dijkstra, as well as the average number of node expansions.}

\label{table:Results}
\centering
\small
\setlength{\tabcolsep}{5pt}
\begin{tabular}{|c | l | *{3}{r} | r |r|}
\toprule
      Range   &    & \multicolumn{3}{c|}{Runtime(ms)} & & \#Exp \\ \cmidrule{3-5}
      kWh & Algorithm  &  \multicolumn{1}{c}{Min.}& \multicolumn{1}{c}{Avg.} & \multicolumn{1}{c|}{Max.} & \multicolumn{1}{c|}{$\eta$} & $\times 10^3$ \\

\midrule
0 - 20     & Dijkstra& 1.9  & 27.6  & 152.8  & 1.00     & 115 \\
      & A*& 1.3  & 15.6  & 157.5  & 2.12 & 42  \\
      & Pr-A*\textsubscript{fw}   & 2.9  & 20.6  & 264.9  & 1.51 & 44  \\
      & Pr-A*\textsubscript{bw}   & 0.8  & 19.6  & 262.6  & 2.03 & 46  \\
      & Pr-BA*& 5.3  & 43.8  & 516.5  & 0.74 & 62  \\
      & Pr-BA*\textsubscript{par} & 2.4  & 30.9  & 509.0    & 1.32 & 62  \\
\midrule
20 - 40 & Dijkstra& 10.1 & 79.5  & 353.6  & 1.00     & 335 \\
      & A*& 4.1  & 44.0    & 184.6  & 1.98 & 122 \\
      & Pr-A*\textsubscript{fw}   & 6.8  & 55.3  & 290.8  & 1.63  & 125 \\
      & Pr-A*\textsubscript{bw}   & 3.4  & 53.1  & 288.8  & 1.91 & 121 \\
      & Pr-BA*& 14.2 & 110.9 & 570.4  & 0.80 & 169 \\
      & Pr-BA*\textsubscript{par} & 6.0    & 75.7  & 543.1  & 1.34 & 167 \\
\midrule
40 - 60 & Dijkstra& 14.1 & 160.7 & 686.8  & 1.00     & 645 \\
      & A*& 5.5  & 79.8  & 327.9  & 2.14 & 214 \\
      & Pr-A*\textsubscript{fw}   & 8.1  & 101.4 & 513.6  & 1.77 & 220 \\
      & Pr-A*\textsubscript{bw}   & 6.5  & 101.8 & 501.3  & 1.95  & 222 \\
      & Pr-BA*& 17.7 & 202.6 & 976.7  & 0.86 & 304 \\
      & Pr-BA*\textsubscript{par} & 10.0   & 139.1 & 942.0    & 1.44 & 302 \\
\midrule
60 - 85 & Dijkstra& 16.3 & 242.9 & 1058.2 & 1.00     & 992 \\
      & A*& 10.0   & 121.7 & 406.8  & 2.05 & 368 \\
      & Pr-A*\textsubscript{fw}   & 12.3 & 151.2 & 549.9  & 1.69 & 378 \\
      & Pr-A*\textsubscript{bw}   & 12.7 & 152.8 & 646.6  & 1.80   & 382 \\
      & Pr-BA*& 27.6 & 298.7 & 1182.2 & 0.83 & 522 \\
      & Pr-BA*\textsubscript{par} & 14.4 & 191.9 & 1060.1 & 1.39 & 516 \\

\bottomrule
\end{tabular}

\end{table}

We report the results for all feasible instances in Table~\ref{table:Results}, where the instances are grouped into four energy ranges based on their minimum computed energy costs, spanning from 0 to 85\,kWh, with at least 240 instances in each group.
The results include the minimum, average (arithmetic), and maximum runtimes (in milliseconds) observed for each algorithm across the instances of the group. The reported runtime includes the search time as well as the initialization time of all data structures and priority queue. We also report the average speedup relative to Dijkstra ($\eta$), and also the average number of node expansions (last column).

\textbf{Runtime analysis:} From the results, we observe that energy-optimal A* search consistently performs approximately twice as fast as Dijkstra across all instance groups. 
This highlights the effectiveness of heuristic-guided search, as the slight overhead from heuristic computation is offset by a significant reduction in search time.
Comparing the profile-based search methods, it is evident that the forward profile search variant, Pr-A*\textsubscript{fw}, performs closely to A*. This is likely because their search spaces are largely overlapping, whereas the backward variant operates over a different search space. That said, the backward profile variant, Pr-A*\textsubscript{bw}, achieves comparable average runtimes to its forward counterpart and is slightly faster in the lower energy range.
Compared to A* with known $\mathcal{E}_{\mathit{init}}$, both Pr-A*\textsubscript{fw} and Pr-A*\textsubscript{bw} are at most 30\% slower, demonstrating the efficiency of our proposed technique and its pseudo-polynomial performance on large-scale networks.
Surprisingly, both bidirectional variants perform worse than the unidirectional profile search. In particular, Pr-BA* is even slower than Dijkstra. A key reason for this is the bidirectional path matching within the fully overlapping search spaces, where the forward and backward frontiers are continuously joined and later evaluated for optimality and potential upper-bound updates, introducing significant computational overhead.
The parallel variant, Pr-BA*\textsubscript{par}, achieves partial performance improvement by enabling concurrent exploration of the forward and backward frontiers.
However, it still remains up to 50\% slower than the unidirectional variants, highlighting the need for more efficient bidirectional energy profile searches.

\begin{figure}[t]
\footnotesize
\centering 
 \includegraphics[width=1\columnwidth]{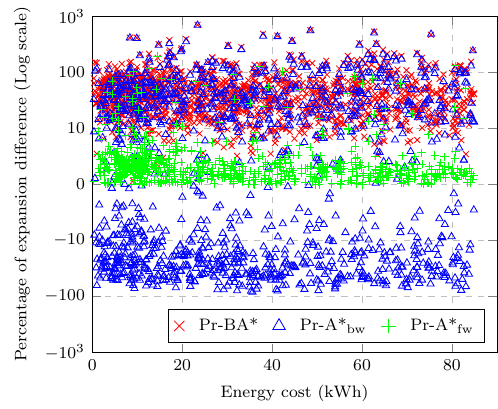}
 \caption{Distribution of expansion differences (\%) for profile-search algorithms relative to energy-optimal A*.}
\label{fig:expansion}
\end{figure}

\textbf{Analysis of node expansion:}
Comparing the number of node expansions, we observe that our forward profile search, Pr-A*\textsubscript{fw}, expands nearly the same number of nodes as the energy-optimal A* search, while the bidirectional approaches expand approximately 50\% more nodes on average. 
To better understand the distribution of expansions, Figure~\ref{fig:expansion} presents the relative difference in node expansions with respect to energy-optimal A* for the three profile-based variants over all feasible instances. The horizontal axis denotes the optimal energy requirement of the instances ($g$-values). We observe that the forward variant, Pr-A*\textsubscript{fw}, incurs less than 10\% extra expansions across most instances. In contrast, the backward variant shows far greater variability: depending on the instance, it can require more than 100\% additional expansions, or conversely, up to 100\% fewer expansions (negative ratios). The bidirectional variant, however, consistently exhibits higher expansion overhead, often approaching or exceeding 100\% additional node expansions.
In terms of the number of solution paths returned, the maximum observed was four, with the vast majority of instances yielding only a single energy-optimal solution path.

\textbf{Comparison to existing solutions:}
We first compare our algorithm with the label-correcting approach of~\citet{SchonfelderLW14}. Their profile A* search is approximately an order of magnitude slower than conventional energy-optimal A*, whereas our approach is less than 30\% slower on average.
In comparison to the profile search method of~\citet{DBLP:journals/algorithmica/BaumDPSWZ20}, we were unable to reproduce their results or find a publicly available, efficient implementation of their energy profile search method. Therefore, we compare their reported performance against our benchmarks indirectly. Their method is a label-correcting approach based on Dijkstra, which similarly uses a height-induced potential function to obtain reduced energy costs for an EV with 85\,kWh battery. According to their reported results, their profile Dijkstra search is up to 50\% slower than the standard energy-optimal Dijkstra.
Given that our energy profile A* search is up to 70\% faster than Dijkstra, we conclude that our approach can be at least twice as fast as the profile search method of~\citet{DBLP:journals/algorithmica/BaumDPSWZ20}, while also being structurally simpler and easier to implement.

\section{Conclusion}
This paper presents a novel A*-based algorithm for energy-optimal profile search for electric vehicles, addressing the practical challenge of energy-efficient path planning under unknown initial energy levels. Unlike existing methods that rely on complex label-correcting strategies, our approach leverages multi-objective search and adopts a straightforward label-setting framework with efficient dominance pruning to propagate energy profiles during the search more effectively.
We implemented both unidirectional and bidirectional variants of the framework and evaluated them on realistic, large-scale road networks enriched with energy consumption data. Experimental results show that the unidirectional profile search variants perform comparably to the energy-optimal A* with known initial energy, while fully supporting energy profile queries. Our approach outperforms existing methods in both simplicity and scalability, offering a practical solution for energy-optimal path planning for EVs in real-world scenarios.

\section{Acknowledgments}
Research supported by the Department of Climate Change, Energy, the Environment and Water under the International Clean Innovation Researcher Networks program grant number ICIRN000077.
Mahdi Jalili is supported by Australian Research Council through projects DP240100963, DP240100830, LP230100439 and IM240100042.

\bibliography{References.bib}

@inproceedings{AhmadiTHK21_cp,
  author    = {Saman Ahmadi and
               Guido Tack and
               Daniel Harabor and
               Philip Kilby},
  editor    = {Laurent D. Michel},
  title     = {Vehicle Dynamics in Pickup-And-Delivery Problems Using Electric Vehicles},
  booktitle = {27th International Conference on Principles and Practice of Constraint
               Programming, {CP} 2021, Montpellier, France (Virtual Conference),
               October 25-29, 2021},
  series    = {LIPIcs},
  volume    = {210},
  pages     = {11:1--11:17},
  publisher = {Schloss Dagstuhl - Leibniz-Zentrum f{\"{u}}r Informatik},
  year      = {2021},
  url       = {https://doi.org/10.4230/LIPIcs.CP.2021.11},
  doi       = {10.4230/LIPIcs.CP.2021.11},
  bibsource = {dblp computer science bibliography, https://dblp.org}
}

@article{dijkstra1959note,
  author    = {Edsger W. Dijkstra},
  title     = {A note on two problems in connexion with graphs},
  journal   = {Numerische Mathematik},
  volume    = {1},
  pages     = {269--271},
  year      = {1959},
  url       = {https://doi.org/10.1007/BF01386390},
  doi       = {10.1007/BF01386390},
  bibsource = {dblp computer science bibliography, https://dblp.org}
}

@article{hart1968formal,
  author    = {Peter E. Hart and
               Nils J. Nilsson and
               Bertram Raphael},
  title     = {A Formal Basis for the Heuristic Determination of Minimum Cost Paths},
  journal   = {{IEEE} Trans. Syst. Sci. Cybern.},
  volume    = {4},
  number    = {2},
  pages     = {100--107},
  year      = {1968},
  url       = {https://doi.org/10.1109/TSSC.1968.300136},
  doi       = {10.1109/TSSC.1968.300136},
  bibsource = {dblp computer science bibliography, https://dblp.org}
}

@article{bellman1958routing,
  title={On a routing problem},
  author={Bellman, Richard},
  journal={Quarterly of applied mathematics},
  volume={16},
  number={1},
  pages={87--90},
  year={1958}
}

@techreport{ford1956network,
  title={Network flow theory},
  author={Ford Jr, Lester R},
  year={1956},
  institution={Rand Corp Santa Monica Ca}
}

@inproceedings{AhmadiSHJ24,
  author       = {Saman Ahmadi and
                  Nathan R. Sturtevant and
                  Daniel Harabor and
                  Mahdi Jalili},
  editor       = {Sara Bernardini and
                  Christian Muise},
  title        = {Exact Multi-objective Path Finding with Negative Weights},
  booktitle    = {Proceedings of the Thirty-Fourth International Conference on Automated
                  Planning and Scheduling, {ICAPS} 2024, Banff, Alberta, Canada, June
                  1-6, 2024},
  pages        = {11--19},
  publisher    = {{AAAI} Press},
  year         = {2024},
  url          = {https://doi.org/10.1609/icaps.v34i1.31455},
  doi          = {10.1609/ICAPS.V34I1.31455},
  bibsource    = {dblp computer science bibliography, https://dblp.org}
}

@article{johnson1977efficient,
  author    = {Donald B. Johnson},
  title     = {Efficient Algorithms for Shortest Paths in Sparse Networks},
  journal   = {J. {ACM}},
  volume    = {24},
  number    = {1},
  pages     = {1--13},
  year      = {1977},
  url       = {https://doi.org/10.1145/321992.321993},
  doi       = {10.1145/321992.321993},
  bibsource = {dblp computer science bibliography, https://dblp.org}
}

@inproceedings{eisner2011optimal,
  author    = {Jochen Eisner and
               Stefan Funke and
               Sabine Storandt},
  editor    = {Wolfram Burgard and
               Dan Roth},
  title     = {Optimal Route Planning for Electric Vehicles in Large Networks},
  booktitle = {Proceedings of the Twenty-Fifth {AAAI} Conference on Artificial Intelligence,
               {AAAI} 2011, San Francisco, California, USA, August 7-11, 2011},
  publisher = {{AAAI} Press},
  year      = {2011},
  url       = {http://www.aaai.org/ocs/index.php/AAAI/AAAI11/paper/view/3637},
  bibsource = {dblp computer science bibliography, https://dblp.org}
}

@inproceedings{baum2013energy,
  author    = {Moritz Baum and
               Julian Dibbelt and
               Thomas Pajor and
               Dorothea Wagner},
  editor    = {Craig A. Knoblock and
               Markus Schneider and
               Peer Kr{\"{o}}ger and
               John Krumm and
               Peter Widmayer},
  title     = {Energy-optimal routes for electric vehicles},
  booktitle = {21st {SIGSPATIAL} International Conference on Advances in Geographic
               Information Systems, {SIGSPATIAL} 2013, Orlando, FL, USA, November
               5-8, 2013},
  pages     = {54--63},
  publisher = {{ACM}},
  year      = {2013},
  url       = {https://doi.org/10.1145/2525314.2525361},
  doi       = {10.1145/2525314.2525361},
  bibsource = {dblp computer science bibliography, https://dblp.org}
}

@inproceedings{storandt2012quick,
  author    = {Sabine Storandt},
  editor    = {Stephan Winter and
               Matthias M{\"{u}}ller{-}Hannemann},
  title     = {Quick and energy-efficient routes: computing constrained shortest paths for electric vehicles},
  booktitle = {5th {ACM} {SIGSPATIAL} International Workshop on Computational Transportation
               Science 2011, CTS'12, November 6, 2012, Redondo Beach, CA, {USA}},
  pages     = {20--25},
  publisher = {{ACM}},
  year      = {2012},
  url       = {https://doi.org/10.1145/2442942.2442947},
  doi       = {10.1145/2442942.2442947},
  bibsource = {dblp computer science bibliography, https://dblp.org}
}

@article{DBLP:journals/transci/BaumDGWZ19,
  author    = {Moritz Baum and
               Julian Dibbelt and
               Andreas Gemsa and
               Dorothea Wagner and
               Tobias Z{\"{u}}ndorf},
  title     = {Shortest Feasible Paths with Charging Stops for Battery Electric Vehicles},
  journal   = {Transportation Science},
  volume    = {53},
  number    = {6},
  pages     = {1627--1655},
  year      = {2019},
  url       = {https://doi.org/10.1287/trsc.2018.0889},
  doi       = {10.1287/trsc.2018.0889},
  bibsource = {dblp computer science bibliography, https://dblp.org}
}

@inproceedings{artmeier2010shortest,
  author    = {Andreas Artmeier and
               Julian Haselmayr and
               Martin Leucker and
               Martin Sachenbacher},
  editor    = {R{\"{u}}diger Dillmann and
               J{\"{u}}rgen Beyerer and
               Uwe D. Hanebeck and
               Tanja Schultz},
  title     = {The Shortest Path Problem Revisited: Optimal Routing for Electric
               Vehicles},
  booktitle = {{KI} 2010: Advances in Artificial Intelligence, 33rd Annual German
               Conference on AI, Karlsruhe, Germany, September 21-24, 2010. Proceedings},
  series    = {Lecture Notes in Computer Science},
  volume    = {6359},
  pages     = {309--316},
  publisher = {Springer},
  year      = {2010},
  url       = {https://doi.org/10.1007/978-3-642-16111-7\_35},
  doi       = {10.1007/978-3-642-16111-7\_35},
  bibsource = {dblp computer science bibliography, https://dblp.org}
}

@book{pearl1984heuristics,
  author    = {Judea Pearl},
  title     = {Heuristics - intelligent search strategies for computer problem solving},
  series    = {Addison-Wesley series in artificial intelligence},
  publisher = {Addison-Wesley},
  year      = {1984},
  isbn      = {978-0-201-05594-8},
  bibsource = {dblp computer science bibliography, https://dblp.org}
}

@inproceedings{geisberger2008contraction,
  author    = {Robert Geisberger and
               Peter Sanders and
               Dominik Schultes and
               Daniel Delling},
  editor    = {Catherine C. McGeoch},
  title     = {Contraction Hierarchies: Faster and Simpler Hierarchical Routing in
               Road Networks},
  booktitle = {Experimental Algorithms, 7th International Workshop, {WEA} 2008, Provincetown,
               MA, USA, May 30-June 1, 2008, Proceedings},
  series    = {Lecture Notes in Computer Science},
  volume    = {5038},
  pages     = {319--333},
  publisher = {Springer},
  year      = {2008},
  url       = {https://doi.org/10.1007/978-3-540-68552-4\_24},
  doi       = {10.1007/978-3-540-68552-4\_24},
  bibsource = {dblp computer science bibliography, https://dblp.org}
}

@article{DBLP:journals/algorithmica/BaumDPSWZ20,
  author    = {Moritz Baum and
               Julian Dibbelt and
               Thomas Pajor and
               Jonas Sauer and
               Dorothea Wagner and
               Tobias Z{\"{u}}ndorf},
  title     = {Energy-Optimal Routes for Battery Electric Vehicles},
  journal   = {Algorithmica},
  volume    = {82},
  number    = {5},
  pages     = {1490--1546},
  year      = {2020},
  url       = {https://doi.org/10.1007/s00453-019-00655-9},
  doi       = {10.1007/s00453-019-00655-9},
  bibsource = {dblp computer science bibliography, https://dblp.org}
}

@inproceedings{SachenbacherLAH11,
  author    = {Martin Sachenbacher and
               Martin Leucker and
               Andreas Artmeier and
               Julian Haselmayr},
  editor    = {Wolfram Burgard and
               Dan Roth},
  title     = {Efficient Energy-Optimal Routing for Electric Vehicles},
  booktitle = {Proceedings of the Twenty-Fifth {AAAI} Conference on Artificial Intelligence,
               {AAAI} 2011, San Francisco, California, USA, August 7-11, 2011},
  publisher = {{AAAI} Press},
  year      = {2011},
  url       = {http://www.aaai.org/ocs/index.php/AAAI/AAAI11/paper/view/3735},
  bibsource = {dblp computer science bibliography, https://dblp.org}
}

@article{DellingGPW17,
  author    = {Daniel Delling and
               Andrew V. Goldberg and
               Thomas Pajor and
               Renato F. Werneck},
  title     = {Customizable Route Planning in Road Networks},
  journal   = {Transp. Sci.},
  volume    = {51},
  number    = {2},
  pages     = {566--591},
  year      = {2017},
  url       = {https://doi.org/10.1287/trsc.2014.0579},
  doi       = {10.1287/trsc.2014.0579},
  bibsource = {dblp computer science bibliography, https://dblp.org}
}

@article{JungP02,
  author    = {Sungwon Jung and
               Sakti Pramanik},
  title     = {An Efficient Path Computation Model for Hierarchically Structured
               Topographical Road Maps},
  journal   = {{IEEE} Trans. Knowl. Data Eng.},
  volume    = {14},
  number    = {5},
  pages     = {1029--1046},
  year      = {2002},
  url       = {https://doi.org/10.1109/TKDE.2002.1033772},
  doi       = {10.1109/TKDE.2002.1033772},
  bibsource = {dblp computer science bibliography, https://dblp.org}
}

@inproceedings{SchonfelderLW14,
  author    = {Ren{\'{e}} Sch{\"{o}}nfelder and
               Martin Leucker and
               Sebastian Walther},
  editor    = {Ching{-}Hsien Robert Hsu and
               Shangguang Wang},
  title     = {Efficient Profile Routing for Electric Vehicles},
  booktitle = {Internet of Vehicles - Technologies and Services - First International
               Conference, IOV, Beijing, China, September 1-3, 2014. Proceedings},
  series    = {Lecture Notes in Computer Science},
  volume    = {8662},
  pages     = {21--30},
  publisher = {Springer},
  year      = {2014},
  url       = {https://doi.org/10.1007/978-3-319-11167-4\_3},
  doi       = {10.1007/978-3-319-11167-4\_3},
  bibsource = {dblp computer science bibliography, https://dblp.org}
}

@misc{ahmadi2024realtime,
      title={Real-Time Energy-Optimal Path Planning for Electric Vehicles}, 
      author={Saman Ahmadi and Guido Tack and Daniel Harabor and Philip Kilby and Mahdi Jalili},
      year={2024},
      eprint={2411.12964},
      archivePrefix={arXiv},
      primaryClass={cs.AI},
      url={https://arxiv.org/abs/2411.12964}, 
}

@article{StewartW91,
  author       = {Bradley S. Stewart and
                  Chelsea C. {White III}},
  title        = {Multiobjective A*},
  journal      = {J. {ACM}},
  volume       = {38},
  number       = {4},
  pages        = {775--814},
  year         = {1991},
  url          = {https://doi.org/10.1145/115234.115368},
  doi          = {10.1145/115234.115368},
  bibsource    = {dblp computer science bibliography, https://dblp.org}
}

@inproceedings{AhmadiRJ25_ESA25,
  author       = {Saman Ahmadi and
                  Andrea Raith and
                  Mahdi Jalili},
  editor       = {Anne Benoit and
                  Haim Kaplan and
                  Sebastian Wild and
                  Grzegorz Herman},
  title        = {A Fast and Simple Algorithm for the Resource Constrained Shortest
                  Path Problem},
  booktitle    = {33rd Annual European Symposium on Algorithms, {ESA} 2025, September
                  15-17, 2025, Warsaw, Poland},
  series       = {LIPIcs},
  volume       = {351},
  pages        = {97:1--97:15},
  publisher    = {Schloss Dagstuhl - Leibniz-Zentrum f{\"{u}}r Informatik},
  year         = {2025},
  url          = {https://doi.org/10.4230/LIPIcs.ESA.2025.97},
  doi          = {10.4230/LIPICS.ESA.2025.97},
  bibsource    = {dblp computer science bibliography, https://dblp.org}
}

@inproceedings{ahmadi_ICAPS25,
  title={Parallelizing multi-objective {A*} search},
  author={Ahmadi, Saman and Sturtevant, Nathan R and Raith, Andrea and Harabor, Daniel and Jalili, Mahdi},
  booktitle={Proceedings of the International Conference on Automated Planning and Scheduling},
  volume={35},
  pages={131--139},
  year={2025}
}

@inproceedings{LTMOA3,
  title={Multi-objective search via lazy and efficient dominance checks},
  author={Hern{\'a}ndez, Carlos and Yeoh, William and Baier, Jorge A and Felner, Ariel and Salzman, Oren and Zhang, Han and Chan, Shao-Hung and Koenig, Sven},
  booktitle={Proceedings of the Thirty-Second International Joint Conference on Artificial Intelligence},
  pages={7223--7230},
  year={2023}
}

\clearpage
\appendix
\section*{Technical Appendix}
\subsection*{Heuristics for Energy-Optimal Pathfinding}
A* search in our energy-based pathfinding problem is guided by the energy consumed along subpaths and a heuristic estimate of the incumbent path to the destination. Energy models for electric vehicles (EVs) can be derived either empirically or through physics-based formulation. We present a heuristic function for each model as follows.

\paragraph{Graphs with actual energy as edge cost.}
We use a basic law from physics to define a model-independent reduced cost as:
\begin{equation} \label{eq:ener_red_pot}
\mathit{cost}_{\mathit{red}}(u,v) = \mathit{cost}(u,v) - \mathit{cost}_{\mathit{pot}}(u,v)
\end{equation}
where the potential energy term is given by:
\begin{equation} \label{eq:ener_pot}
\mathit{cost}_{\mathit{pot}}(u,v) = Ma \cdot [H(v) - H(u)]
\end{equation}
Here, $M$ is the total mass of the EV, $a$ is the gravitational acceleration (9.8\,m/s\textsuperscript{2}), and $H : S \rightarrow \mathbb{R}$ is a height function derived from elevation data.
We define a heuristic function based on this reduced cost as:
\begin{equation} \label{eq:ener_heur_pot}
h_e(u) = \lambda \hat{\mathcal{E}} \cdot h_d(u) + \mathit{cost}_{\mathit{pot}}(u, \mathit{goal})
\end{equation}
where $h_d : S \rightarrow \mathbb{R}^+$ is a consistent distance-based heuristic (e.g., haversine distance), $\hat{\mathcal{E}}$ is the average energy efficiency of the EV (in Wh/100m), and $\lambda \ge 0$ is a scaling factor. The value of $\lambda$ is constrained such that:
\[
0 \le \lambda \le \frac{\mathit{cost}_{\mathit{red}}(u,v)}{\hat{\mathcal{E}} \cdot d(u,v)}
\]
where $d(u,v)$ is the length of the link $(u,v)$. The maximum valid $\lambda$ is computed via a linear scan over all edges.

\begin{figure}[t]
\centering
\footnotesize
\begin{tikzpicture}[->,>=stealth',shorten >=1pt,thick]
\SetGraphUnit{3} 
\GraphInit[vstyle=Normal] 
\SetVertexNormal[Shape=circle,MinSize=1cm,LineWidth =1pt]
\tikzset{VertexStyle/.append style = {font=\large\bfseries},thick} 
\Vertex[L=$v$]{v} 
\SOWE[L=$u$](v){u} 
\SOEA[L=$\mathit{goal}$](v){goal} 
\draw[->,black] (u) to node [sloped,above] {$cost_{red}(u,v)$} node[sloped,below]{$l_{uv}$} (v);
\draw[dashed][->,black] (u) to node [above] {$h_e(u)$} node[below]{$h_d(u)$} (goal);
\draw[dashed][->,black] (v) to node [sloped,above] {$h_e(v)$} node[sloped,below]{$h_d(v)$} (goal);  
\end{tikzpicture}
\caption{\small A schematic of distance- and energy-based heuristics.}
\label{fig:heuristic_diagram}
\end{figure}
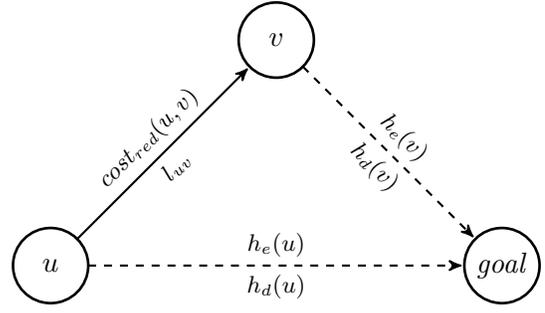

We now show why $h_e$ is consistent and admissible.
Consider edge $(u,v)$ in Figure~\ref{fig:heuristic_diagram}. Based on the natural characteristics of road networks where triangle inequality holds for distances, and given $h_d$ as a consistent distance heuristic, for a given edge $(u,v)$ we have:
\begin{align}
&h_d(u) \le d(u,v) + h_d(v) \Rightarrow h_d(u) - h_d(v) \le d(u,v) \label{eq:ener_pot_cons01} \\
&\Rightarrow \lambda \hat{\mathcal{E}} (h_d(u) - h_d(v)) \le \lambda  \hat{\mathcal{E}} \cdot d(u,v) \label{ener_pot_cons02} \\
&\Rightarrow h_e(u) - h_e(v) \le \lambda \hat{\mathcal{E}} \cdot d(u,v) \label{ener_pot_cons03} \\
&\text{and since } \lambda  \hat{\mathcal{E}} \cdot d(u,v) \le \mathit{cost}_{\mathit{red}}(u,v), \text{and} \label{ener_pot_cons04} \\
&\mathit{cost}_{\mathit{pot}}(u, \mathit{goal}) - \mathit{cost}_{\mathit{pot}}(v, \mathit{goal}) = \mathit{cost}_{\mathit{pot}}(u,v), \\
&\Rightarrow h_e(u) - h_e(v) \le \mathit{cost}(u,v) \label{ener_pot_cons05}
\end{align}
Relation (\ref{ener_pot_cons05}) above denotes $h_e$ is consistent. Given that $h_e(\mathit{goal}) = 0$ and that consistency implies admissibility~\cite{pearl1984heuristics}, $h_e$ is also admissible.

This heuristic remains consistent even under battery constraints (e.g., energy overflow on negative-cost edges). If an EV starts a downhill link with 100\% SoC and cannot store recovered energy, the effective energy cost is adjusted (e.g., from $\mathit{cost}_e < 0$ to $\mathit{cost}'_e = 0$), increasing $\mathit{cost}(u,v)$ temporarily, and thereby preserving the consistency of $h_e$.

\paragraph{Energy models based on longitudinal dynamics.}
We now consider a physics-based energy model adapted from our previous work~\cite{ahmadi2024realtime}, representing energy consumption as a function of road slope and load:
\begin{equation} \label{eq:ener_link}
\mathit{cost} = m(\alpha_2 s^2 + \alpha_1 s + \alpha_0)l + (\beta_2 s^2 + \beta_1 s + \beta_0)l
\end{equation}
where $s = \sin(\theta)$ is the road angle, $l$ is the segment length (in 100m units), and $m$ is the additional EV mass (e.g., passengers weight). Coefficients $\alpha_i, \beta_i$ are vehicle-specific and learned via regression for any particular driving pattern.

For this function, we approximate the coefficients $\alpha_1$ and $\beta_1$ using their average values, $\overline{\alpha_1}$ and $\overline{\beta_1}$ (averaged over all available driving patterns), to derive a \textit{path-independent} potential function, and consequently, a reduced cost defined as:
\begin{equation} \label{eq:ener_pi}
\mathit{cost}_{\mathit{pi}}(u,v) = m(\overline{\alpha_1} + \overline{\beta_1})\cdot[H(v) - H(u)]
\end{equation}
\begin{equation} \label{eq:ener_red_pi}
\mathit{cost}_{\mathit{red}}(u,v) = \mathit{cost}(u,v) - \mathit{cost}_{\mathit{pi}}(u,v)
\end{equation}

We then define the corresponding heuristic as:
\begin{equation} \label{eq:ener_heur_pi}
h_e(u) = \lambda (m{\alpha_0}_{\min} + {\beta_0}_{\min}) h_d(u) + \mathit{cost}_{\mathit{pi}}(u, \mathit{goal})
\end{equation}
Here, ${\alpha_0}_{\min}, {\beta_0}_{\min}$ are the minimum energy coefficients ${\alpha_0}$ and ${\beta_0}$ across patterns, respectively, and $\lambda$ is a scaling factor satisfying:
\[
0 \le \lambda \le \frac{\mathit{cost}_{\mathit{red}}(u,v)}{\mathit{cost}_{\min}(u,v)}
\]
where:
\begin{equation} \label{eq:ener_min}
\mathit{cost}_{\min}(u,v) = (m{\alpha_0}_{\min} + {\beta_0}_{\min}) d(u,v)
\end{equation}

To prove consistency, using triangle inequality:
\begin{equation} \label{eq:ener_pot_cons1}
h_d(u) - h_d(v) \le d(u,v)
\end{equation}
\begin{equation} \label{ener_pot_cons3}
\Rightarrow h_e(u) - h_e(v) \le \lambda \mathit{cost}_{\min}(u,v) + \mathit{cost}_{\mathit{pi}}(u,v)
\end{equation}
\begin{equation} \label{ener_pot_cons4}
\Rightarrow h_e(u) - h_e(v) \le \mathit{cost}(u,v)
\end{equation}

Therefore, this model-specific energy heuristic is also consistent and admissible, and remains valid under energy constraints, as any required adjustment will only increase the energy cost temporarily.

\begin{figure*}[t]
\footnotesize
\centering 
\includegraphics[width=1\textwidth]{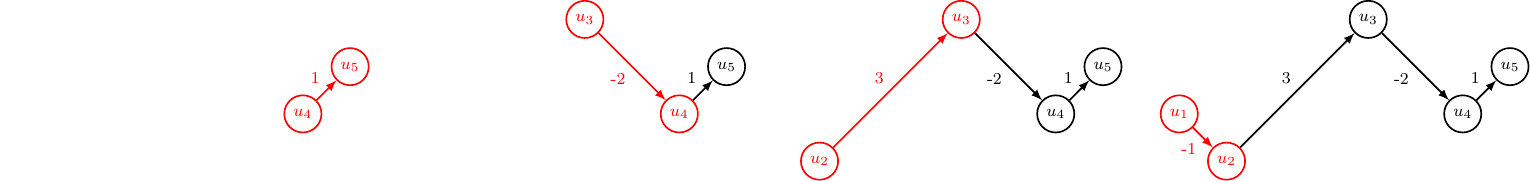} \\
\vspace{0.1cm}
\includegraphics[width=1\textwidth]{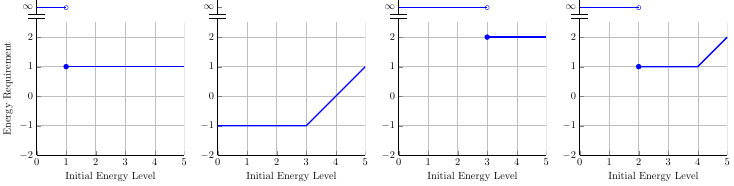}
\caption{\small A schematic of energy profile linking for a simple path. The linking procedure is performed in the backward direction, starting with the last edge (leftmost plot). The final energy profile of the path is shown in the rightmost plot. Maximum energy is set to $\mathcal{E}_{\mathit{max}}=5$.}
\label{linking_profiles}
\end{figure*}

\subsection{Linking Energy Profiles}

To better illustrate the linking procedure, we link the energy profiles of the edges along a simple path shown in Figure~\ref{linking_profiles}. The path consists of four edges (and four corresponding energy profiles), and we perform a backward profile linking. We assume the maximum energy level is $\mathcal{E}_{\mathit{max}} = 5$.

We begin with the last edge, $u_4 \leadsto u_5$, which has an energy cost of 1. Its energy profile is shown in the leftmost plot of Figure~\ref{linking_profiles}, with $\mathcal{E}_{\mathit{min}} = g = \overline{g} = 1$.
Next, we extend the path by adding the preceding edge, $u_3 \leadsto u_4$, with cost -2. Since this link has a negative energy cost, both the minimum required energy $\mathcal{E}_{\mathit{min}}$ and the energy cost $g$ of the subpath decrease by two units. As a result, the minimum cost of the extended path $u_3 \leadsto u_4 \leadsto u_5$ becomes -1, and $\mathcal{E}_{\mathit{min}}$ becomes 0. This corresponds to shifting the initial profile both to the left and downward by two units. However, since the EV cannot recuperate energy when starting from a full battery at $\mathcal{E}_{\mathit{max}}$, the maximum energy cost remains unchanged, i.e., $\overline{g} = 1$, as shown in the second plot of Figure~\ref{linking_profiles}.

In the third step, we extend the subpath by adding the positive-cost edge $u_2 \leadsto u_3$ with a cost of 3. Accordingly, both the energy cost and $\mathcal{E}_{\mathit{min}}$ of the subpath increase. This is equivalent to shifting the profile up and to the right by three units. The resulting minimum energy cost is now 2, and $\mathcal{E}_{\mathit{min}} = 3$. The maximum energy cost, however, becomes equal to the minimum cost, i.e., $\overline{g} = 2$, since no overcharging occurs even when starting from a full battery at $u_2$. The updated profile is shown in the third plot of Figure~\ref{linking_profiles}.

Finally, we add the first edge, $u_1 \leadsto u_2$, with an energy cost of -1. Similar to the earlier negative-cost link, the profile is shifted left and downward by one unit, reducing both $g$ and $\mathcal{E}_{\mathit{min}}$ by one, while keeping the maximum energy cost $\overline{g}$ from the preceding path unchanged. This final energy profile is shown in the rightmost plot of Figure~\ref{linking_profiles}.
Therefore, to traverse the complete path from $u_1$ to $u_5$, at least two units of energy are required at $u_1$, and the total energy cost lies in the range $[1,2]$, depending on the initial energy level.

\subsection{Energy Profile Search with Backward A*}
Algorithm~\ref{alg:Astarbw} presents a pseudocode for our energy-optimal profile search using backward A*. It is nearly identical to the forward variant in Algorithm~\ref{alg:Astarfw}, with two main differences in: i) the calculations of $\mathcal{E}_{\mathit{min}}$ and $\overline{g}$ breakpoints during expansion; ii) the heuristic function $h$, which is now a backward energy heuristic that returns energy estimates relative to the $\mathit{start}$ state. Note that, in the backward search, $\mathcal{E}_{\mathit{min}}$ and $\overline{g}$ denote maximum suffix and prefix, respectively, when extending paths from $\mathit{goal}$ to $\mathit{start}$.
Thus, the upper-bounding at line~\ref{alg3:energy-check2} of Algorithm~\ref{alg:Astarbw} is conducted using $\mathcal{E}_{\mathit{min}}$ (the maximum suffix plus cost-to-go estimate to $\mathit{start}$).

\begin{algorithm}[ht]
\small
\caption{Energy Profile (Backward) A* Search}
\label{alg:Astarbw}
\DontPrintSemicolon
 \KwInput{A problem instance ($G$, $\mathit{cost}$, $h$, $\mathit{start}$, $\mathit{goal}$)}
 \KwOutput{A superset of nodes representing optimal profiles}
 $\mathit{Open} \gets \emptyset$, \ $\overline{f} \gets \infty$ \;
$\mathcal{X}(u) \gets \emptyset$ $\forall u \in S$\;

 $x \gets $ new node with $s(x) = \mathit{goal}$\ \;
 $g(x) \gets 0 $, $\overline{g}(x) \gets 0$, $\mathcal{E}_{\mathit{min}}(x) \gets 0$, $ parent(x) \gets \varnothing$ \;
 $f(x) \gets h(\mathit{goal}) $\;
Add $x$ to $Open$\;
\While{$Open \neq \emptyset$}
{
 Extract from $Open$ a node $x$ with the smallest $f$-value  \;
     \lIf{$f(x) \geq \overline{f}$ }
     { {\bf break} }
     
     
    

     $\mathit{dominated} \gets \mathrm{false}$\label{alg:match:dom1}\;
     \ForEach{$y \in \mathcal{X}(s(x))$\label{alg:nwrca:removedom1}}
        { 
       \If{$\mathcal{E}_{\mathit{min}}(y) \leq \mathcal{E}_{\mathit{min}}(x) $ {\bf and} $\overline{g}(y) \leq \overline{g}(x) $}
            {
            $\mathit{dominated} \gets \mathrm{true}$ \; 
        {\bf break}\label{alg:nwrca:removedom2}} 
        }
     \lIf{$\mathit{dominated} = \mathrm{true}$} {\textbf{continue}}

     add $x$ to $\mathcal{X}(s(x))$\label{alg:nwrca:add}\;
     
     \If{$s(x)=\mathit{start}$}
     { 
     
     $\overline{f} \gets \mathrm{Min} (\overline{f},\mathrm{Max}(\mathcal{E}_{\mathit{min}}(x), \overline{g}(x)))$ \;
     {\bf continue}  \;}

    \ForEach{$v \in Pred(s(x))$}
        { 
        $y \gets $ new node with $s(y) = v$ \;       
        $cost_{e} \gets \mathit{cost}(v, s(x))$ \;
        ${g}(y) \gets \mathrm{Max}({g}(x) + \mathit{cost}_{e}, -\mathcal{E}_{\mathit{max})}$ \;
        $f(y) \gets g(y) + h(v)$ \;
        $\overline{g}(y) \gets \mathrm{Max}(\overline{g}(x), {g}(y)$) \;
        $\mathcal{E}_{\mathit{min}}(y) \gets \mathrm{Max}(0,\mathcal{E}_{\mathit{min}}(x)+cost_{e})$ \;
        $parent(y) \gets x$ \;
        \If{$\mathcal{E}_{\mathit{min}}(y) > \mathcal{E}_{\mathit{max}}$ {\bf or} $\overline{g}(y) > \mathcal{E}_{\mathit{max}}$ \label{alg3:energy-check1}}
            { {\bf continue}}
        \lIf{$\mathcal{E}_{\mathit{min}}(y) + h(v) > \mathcal{E}_{\mathit{max}}$ \label{alg3:energy-check2}}
            { {\bf continue}}

        $z \gets $ last node in $\mathcal{X}(v)$ \;
        \If{$\mathcal{E}_{\mathit{min}}(z) \leq \mathcal{E}_{\mathit{min}}(y) $ {\bf and} $\overline{g}(z) \leq \overline{g}(y) $}
            { {\bf continue}}

        Add $y$ to $Open$ \label{alg3:relax-check2}\;
        
        }
       
}
\Return{$\mathcal{X}(\mathit{start})$}\; 
\end{algorithm}

\subsection{Energy Profile Search with Bidirectional A*}
Algorithm~\ref{alg:Astarbidir} presents a pseudocode for energy-optimal profile search using bidirectional A*. This framework maintains two priority queues (one for forward and another for backward search), as well as separate heuristic functions for each direction. Each priority queue is initialized with a corresponding initial node ($\mathit{start}$ in the forward direction and $\mathit{goal}$ in the backward direction).
Each iteration of the algorithm involves extracting the best node from either of the queues. Each node in this framework also stores a tag, called $\mathit{joined}$, denoting solution nodes obtained by matching a forward node with a backward counterpart (both with the same state).
When a joined node is extracted, the algorithm checks it against all existing solution profiles stored in the solution set $\mathit{Sols}$ for dominance, and if deemed non-dominated, it will be added to the solution set.
Otherwise, if the extracted node is not a joined solution node, it will get expanded as in forward or backward A* search, depending on the priority queue from which the node is extracted.
The last step is joining the non-dominated node with all nodes expanded with the same state in the opposite direction to form a complete (joined) path.
Upon joining a forward node with a backward node, we generate a new profile whose breakpoints are calculated through linking the corresponding forward and backward profiles. In other words, the resulting joined profile is still represented by two breakpoints, which can be done in constant time (see our implementation for more details). Once the joined node is set up, it will be added to the corresponding priority queue. 
The search eventually terminates and returns $\mathit{Sols}$ as a set containing all joined solution nodes representing optimal energy profiles.

\begin{algorithm}[!t]
\small
\caption{Bidirectional Energy Profile A*}

\label{alg:Astarbidir}
\DontPrintSemicolon
 \KwInput{A problem instance ($G$, $\mathit{cost}$ , $h$, $\mathit{start}$, $\mathit{goal}$)}
 \KwOutput{A superset of node pairs representing optimal profiles}

$\mathit{Sols} \gets \emptyset $\;
$\mathit{Open}^f \gets \emptyset, \mathit{Open}^b \gets \emptyset $ \;
$\mathcal{X}^f(u) \gets \emptyset, \mathcal{X}^b(u) \gets \emptyset$ $\forall u \in S$ \;
$x \gets $ new node with $s(x) = \mathit{start}$\ \;
$y \gets $ new node with $s(y) = \mathit{goal}$\ \;
 $ {g}(x) \gets 0, {g}(y) \gets {0}$\;
 $ {f}(x) \gets {h}^f(\mathit{start})$,
 $ {f}(y) \gets {h}^b(\mathit{goal})$\;

 $\mathcal{E}_{\mathit{min}}(x) \gets 0$,
  $\mathcal{E}_{\mathit{min}}(y) \gets 0$ \;
$ parent(x) \gets \varnothing$,
 $ parent(y) \gets \varnothing$ \;
add $x$ to $\mathit{Open}^f$ and $y$ to $\mathit{Open}^b$\;
\While{$\mathit{Open}^f \bigcup \mathit{Open}^b \neq \emptyset$}
{
  {extract from $\mathit{Open}^f \bigcup \mathit{Open}^b$ node $x$ with the smallest ${f}$-value \label{alg:HL:least_cost} \;}
  \nlast
  \lIf{$  f(x) \geq \overline{f} $} 
      {\textbf{break}}
 $\mathit{dominated} \gets \mathrm{false}$\;
\If{$\mathit{joined}(x) = \mathit{true}$}
    {
   
    \ForEach{$y \in \mathit{Sols}$}
    { 
       \If{$\mathcal{E}_{\mathit{min}}(y) \leq \mathcal{E}_{\mathit{min}}(x) $ {\bf and} $\overline{g}(y) \leq \overline{g}(x) $}
            {
            $\mathit{dominated} \gets \mathrm{true}$ \; 
        {\bf break}\label{alg:bidir:removedom2}} 
        }
    \If{$\mathit{dominated} = \mathrm{false}$} { 
         $\overline{f} \gets \mathrm{Min} (\overline{f},\mathrm{Max}(\mathcal{E}_{\mathit{min}}(x), \overline{g}(x)))$ \;
         Add $x$ to $\mathit{Sols}$\;}
    \textbf{continue} \;
    
    }
    
\ForEach{$y \in \mathcal{X}(s(x))$\label{alg:bidir:removedom1}}
        { 
       \If{$\mathcal{E}_{\mathit{min}}(y) \leq \mathcal{E}_{\mathit{min}}(x) $ {\bf and} $\overline{g}(y) \leq \overline{g}(x) $}
            {
            $\mathit{dominated} \gets \mathrm{true}$ \; 
        {\bf break}\label{alg:bidir:removedom3}} 
        }
\lIf{$\mathit{dominated} = \mathrm{false}$} { 
    \textbf{continue} }        
{add $x$ to $\mathcal{X}^d(s(x))$}  \;
$d \gets$ direction from which $x$ was extracted\;
 $d' \gets$ opposite direction of $d$ \;

Expand in direction $d$ (see lines 20-33 Algorithms 2,3)\;      

\ForEach{$z \in \mathcal{X}^{d'}(s(x))$}
    {  
    $y \gets $ A new node linking profiles of $x$ and $z$ \;
    $\mathit{parent}(y) \gets (x,z)$ \;
    $\mathit{joined}(y) \gets \mathit{true}$ \;
    add $y$ to $\mathit{Open}^d$\;

    }

}
\Return{$\mathit{Sols}$}
\end{algorithm}

\end{document}